\documentclass[letterpaper]{article} 
\usepackage{aaai2027}  
\usepackage[hyphens]{url}  
\usepackage{graphicx} 
\usepackage{natbib}  
\usepackage{caption} 
\usepackage{amsmath}
\usepackage{amssymb}
\usepackage{algorithm}
\usepackage{algorithmic}
\usepackage{booktabs}
\title{Minimum-Cost Robust Evidence Certificates for Horn Reasoning under Correlated Source Failures}
\author{
    Yang Yukun\textsuperscript{\rm 1}\corresponding
}
\affiliations{
    \textsuperscript{\rm 1}School of Management, Beijing Institute of Technology\\
    yangyukun2005@163.com
}

\begin{document}

\maketitle

\begin{abstract}
A small justification for a rule-based conclusion may disappear when a source is withdrawn, and multiple derivations need not provide independent evidence. We study minimum-cost certificates that preserve a Horn conclusion under every withdrawal of at most $k$ source domains. Evidence records remain distinct from asserted facts, supporting replicated attestations and overlapping failure dependencies. Transversals of minimal-support footprints characterize robustness and explain why disjoint-proof requirements can reject valid certificates. An exact procedure alternates weighted evidence selection with a source-failure oracle. A rule-closed counterinterpretation yields a globally valid selection cut without enumerating supports or failure scenarios. We prove correctness, finite termination, and cost optimality, establish coNP-completeness of unrestricted verification and NP-completeness of synthesis for every fixed budget, and identify a tractable reachability restriction. Experiments on controlled benchmarks, reachability graphs, and ontology-derived instances demonstrate that our counterinterpretation-guided synthesis achieves up to 92\% optimum coverage within practical resource bounds, outperforming scenario expansion and general abduction while confirming the necessity of transversal-based reasoning over proof-packing heuristics.
\end{abstract}


\section{Introduction}

Rule-based decisions often require premises that can be inspected, retained, or communicated. A release policy may combine signed test reports and attestations; a knowledge graph may infer relationships from several institutions' records. Efficient materialization computes conclusions \cite{motik_parallel_2014}, while explanation and provenance expose their dependencies. When a provider becomes unavailable or is removed from an audit, the practical question is whether the retained evidence can still justify the decision.

An inclusion-minimal explanation removes unnecessary premises, and a minimum-cost explanation reduces inspection or disclosure cost. Neither protects against withdrawal of its own premises: one cheap support may concentrate the entire justification in a single supplier. Retaining the full knowledge base can preserve alternatives but exceed an audit budget. We therefore select a fixed evidence package before failures occur and allow its successful derivation to change afterward. The quantifier order is evidence selection, universal source withdrawal, and existence of a surviving justification. Recovery uses only the selected records, as required by an archived audit package; it does not depend on later evidence acquisition.

Counting proofs does not resolve this problem. Different derivations may reuse a record or depend on services within one failure domain. Conversely, requiring domain-disjoint proofs can reject evidence that survives every admissible withdrawal. Robustness depends on whether failed domains intersect all supports, not on whether supports can be packed without overlap.

\textbf{Example 1 (Release authorization).} Authorization requires any two of checks $a,b,c$, whose records depend on domains $s_A,s_B,s_C$. Retaining all three survives any single-domain withdrawal because any two suffice, yet their support footprints $\{s_A,s_B\}$, $\{s_A,s_C\}$, and $\{s_B,s_C\}$ share domains pairwise, precluding disjoint proofs. An alternative attestation $d$ independently authorizes release at higher cost. Optimal robust certification must navigate both inspection costs and failure dependencies.

While why-provenance \cite{buneman_why_2001}, semiring provenance \cite{green_provenance_2007}, and circuit representations \cite{deutch_circuits_2014} expose derivation dependencies, they do not select cost-optimal subsets resilient to source loss. Database resilience \cite{freire_complexity_2015} and Datalog deletion causality \cite{thapa_causality_2026} analyze answer vulnerability to tuple removals via minimal supports, but do not address ex-ante evidence synthesis under grouped, correlated domain withdrawals. Similarly, abduction-based explanations \cite{ignatiev_abduction-based_2019}, robust perturbation models \cite{la_malfa_guaranteed_2021}, and lineage-driven fault injection \cite{alvaro_lineage-driven_2015} seek explanations or fault witnesses without synthesizing minimum-cost packages resilient to source loss. We turn this support--transversal principle into a cost-sensitive separation framework specialized for correlated source domains.

Our synthesis method alternates an exact weighted hitting-set master with an adversarial oracle. A defeating withdrawal yields a rule-closed counterinterpretation and a positive clause that every robust selection must satisfy. This uses established implicit hitting-set separation \cite{saikko_implicit_2016,davies_solving_2011}, with globally valid record cuts for shared failures. Horn closure handles recursive rules without admitting unsupported cycles as proofs.

Our contributions are: (1) a record-level certification objective capturing correlated source failures; (2) a footprint characterization proving why proof packing is incomplete; (3) an exact synthesis algorithm with finite termination and cost optimality guarantees; (4) complexity results showing coNP-complete verification and NP-complete synthesis for fixed budgets; (5) a polynomial flow reduction for independent reachability; and (6) empirical evaluations demonstrating up to 92\% optimum coverage and the necessity of cut strengthening over proof packing.

\section{Preliminaries}

\subsection{Horn consequences and supports}

Let $V$ be a finite universe of ground atoms and $P$ a finite definite Horn program. Each rule $r$ has a head $\operatorname{head}(r)\in V$ and a finite body $\operatorname{body}(r)\subseteq V$; empty bodies are unconditional. Negation, priorities, and integrity constraints are excluded. For facts $X$, least-fixed-point consequences \cite{van_emden_semantics_1976} are

\begin{equation}
\begin{aligned}
X_0 &= X,\\
X_{i+1} &= X_i\cup\{\operatorname{head}(r):r\in P,\ \operatorname{body}(r)\subseteq X_i\},\\
\operatorname{cl}_P(X) &= \bigcup_{i\geq 0}X_i.
\end{aligned}
\label{eq:closure}
\end{equation}

At most $|V|$ strict growth steps occur. Rule counters compute closure in linear time in the explicit ground representation \cite{dowling_linear-time_1984}, which is the input model for all complexity results. First-order grounding is a separate cost.

\textbf{Definition 1 (Support).} Given trusted facts $B\subseteq V$, selectable records $\mathcal E$, and a mapping $\alpha:\mathcal E\rightarrow V$, a support for $q$ is a record set $S\subseteq\mathcal E$ satisfying $q\in\operatorname{cl}_P(B\cup\alpha(S))$. It is minimal if no proper subset is a support. Here $\alpha(S)=\{\alpha(e):e\in S\}$.

Distinct records asserting the same atom maintain separate identities and failure dependencies, generalizing assumption-based truth maintenance \cite{de_kleer_assumption-based_1986}. While rule-derivation trees reconstruct acyclic proofs for recursive programs, minimal supports are defined by semantic entailment, distinct from cardinality or cost minimality. For a family $\mathcal{A}$, $\operatorname{Min}_{\subseteq}(\mathcal{A})$ removes non-minimal supersets.

\subsection{Relation to provenance and abduction}

Unlike recursive semiring provenance \cite{bourgaux_revisiting_2022} or why-provenance enumeration \cite{calautti_complexity_2024,calautti_computing_2024}, which face exponential compilation blowup \cite{darwiche_knowledge_2002}, we target polynomial Boolean entailment under retained subsets. Query causality \cite{meliou_complexity_2010} and database resilience \cite{freire_complexity_2015} study individual tuple deletions rather than adversarial domain withdrawals. While Horn abduction \cite{kakas_abductive_1992}, sufficient reasons \cite{darwiche_reasons_2020}, and diagnosis \cite{reiter_theory_1987} select static premises, they do not guarantee survival against subsequent failures; and unlike minimum label cuts \cite{zhang_minimum_2020}, Horn conjunctions preclude graph reductions.

\section{Robust Evidence Certificates}

\subsection{Records, domains, and the certification objective}

\textbf{Definition 2 (Certification instance).} A certification instance is a tuple $\mathcal I=(V,P,B,\mathcal E,D,\alpha,\gamma,c,q,k)$. The first three components are the ground vocabulary, rules, and trusted facts. The finite sets $\mathcal E$ and $D$ contain evidence records and failure domains. Each record $e$ asserts $\alpha(e)\in V$, has dependencies $\gamma(e)\subseteq D$, and incurs strictly positive rational cost $c(e)$. The target is $q\in V$, and $k$ is a nonnegative integer with $k\leq |D|$.

A domain represents a common withdrawal cause, such as a publisher or hosting service. Dependencies may overlap: failure of any domain in $\gamma(e)$ invalidates $e$. An empty dependency set protects a record. Costs are paid once per retained record; rules and trusted facts remain fixed.

For a selected certificate $C\subseteq\mathcal E$ and failed domains $F\subseteq D$, define the surviving records and asserted facts by

\begin{equation}
\begin{aligned}
\operatorname{surv}(C,F)&=\{e\in C:\gamma(e)\cap F=\varnothing\},\\
X(C,F)&=B\cup\alpha(\operatorname{surv}(C,F)).
\end{aligned}
\label{eq:surv}
\end{equation}

\textbf{Definition 3 (Robust certificate).} A record set $C$ is a $k$-robust certificate for $q$ if

\begin{equation}
\forall F\subseteq D,\quad |F|\leq k\ \Longrightarrow\ q\in\operatorname{cl}_P(X(C,F)).
\label{eq:robust}
\end{equation}

The synthesis task returns a feasible minimizer, or reports that none exists:

\begin{equation}
\begin{aligned}
&\min_{C\subseteq\mathcal E}\operatorname{cost}(C),\qquad
\operatorname{cost}(C)=\sum_{e\in C}c(e),\\
&\text{subject to Equation~\eqref{eq:robust}.}
\end{aligned}
\label{eq:synthesis}
\end{equation}

The guarantee is formal entailment relative to a trust boundary, not objective truth. Withdrawal removes assertions without adding negations; conflicting reports require a separate semantics.

\textbf{Example 2.} Continuing Example~1, let $B=\varnothing$ with rules $q\leftarrow a,b$; $q\leftarrow a,c$; $q\leftarrow b,c$; and $q\leftarrow d$. Records $e_A,e_B,e_C$ assert $a,b,c$ with domains $\{s_A\},\{s_B\},\{s_C\}$ and unit cost; $e_D$ asserts $d$ with $\{s_D\}$ at cost 3. For $k=0$, any pair of checks is optimal (cost 2). For $k=1$, $\{e_A,e_B,e_C\}$ is optimal (cost 3), surviving any single-domain loss. For $k=2$, all four records are required (cost 5), since withdrawing $s_D$ plus one check leaves two checks, while withdrawing two check domains leaves $d$. Replicating $a$ under $\{s_A\}$ provides no resilience, whereas a distinct domain expands feasible certificates.

\subsection{A source-level characterization}

Let $\mathcal S_q(C)$ be the inclusion-minimal supports contained in $C$. For each such support, its footprint is the union of its records' dependencies. We retain only inclusion-minimal footprints:

\begin{equation}
\begin{aligned}
\lambda(S)&=\bigcup_{e\in S}\gamma(e),\\
\mathcal G_q(C)&=\operatorname{Min}_{\subseteq}\{\lambda(S):S\in\mathcal S_q(C)\}.
\end{aligned}
\label{eq:footprint}
\end{equation}

A transversal intersects every member of a family. Define $\tau(\mathcal G)$ as the smallest transversal cardinality. The empty family has $\tau=0$, since no support exists; a family containing the empty set has $\tau=\infty$, since that support cannot be intersected. Define the failure radius

\begin{equation}
\rho_q(C)=\min\{|F|:F\subseteq D,\ q\notin\operatorname{cl}_P(X(C,F))\},
\label{eq:radius}
\end{equation}

with value $\infty$ when no withdrawal defeats the target.

\textbf{Theorem 1 (Grouped support--cut equivalence).} For every certificate candidate $C$, $\rho_q(C)=\tau(\mathcal G_q(C))$. Consequently, $C$ is $k$-robust exactly when $\tau(\mathcal G_q(C))>k$.

\textit{Proof.} Entailment $q\in\operatorname{cl}_P(X(C,F))$ holds iff some minimal support $S\in\mathcal S_q(C)$ survives $F$, which requires $F\cap\lambda(S)=\varnothing$. Thus, $F$ defeats $C$ ($q\notin\operatorname{cl}_P(X(C,F))$) iff $F$ intersects every support footprint in $\mathcal G_q(C)$, i.e., $F$ is a transversal of $\mathcal G_q(C)$. Minimizing $|F|$ yields $\rho_q(C)=\tau(\mathcal G_q(C))$. Therefore, $C$ is $k$-robust iff no defeating withdrawal has size $\le k$, which is precisely $\tau(\mathcal G_q(C))>k$. $\square$

\textbf{Proposition 1 (Disjointness is sufficient but incomplete).} Having $k+1$ supports with pairwise disjoint nonempty footprints implies $k$-robustness. The converse fails already for $k=1$.

\textit{Proof.} A failed domain can intersect at most one of the disjoint footprints, so $k$ failed domains cannot intersect all $k+1$. In Example 2, the check certificate has footprints $\{s_A,s_B\}$, $\{s_A,s_C\}$, and $\{s_B,s_C\}$. Every transversal needs two domains, yet no two footprints are disjoint. $\square$

\begin{figure*}[t]
\centering
\includegraphics[width=0.735\textwidth]{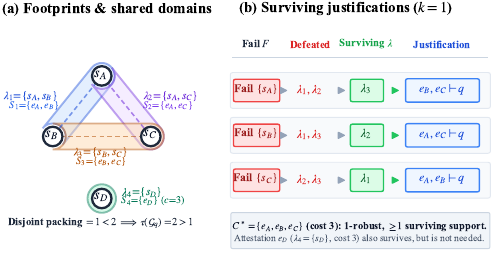}
\caption{Running-example support footprints, their shared domains, and a surviving justification for each admissible single-domain withdrawal.}
\label{fig:footprints}
\end{figure*}

Figure~\ref{fig:footprints} illustrates the running-example footprints and surviving supports. Footprints suffice for verification of a fixed certificate, but optimization still needs record identities: supports with the same footprint may have different costs, and shared records are charged once.

\subsection{Monotonicity and explanatory necessity}

\textbf{Proposition 2 (Feasibility and nominal reduction).} Robustness is upward closed in $C$ and downward closed in $k$. A feasible certificate exists exactly when $\mathcal E$ is robust. At $k=0$, minimum-cost certification coincides with minimum-cost positive Horn support selection.

\textit{Proof.} Closure monotonicity ensures expanding $C$ preserves surviving derivations under any $F$, while decreasing $k$ restricts withdrawals. At $k=0$, $F=\varnothing$, reducing Equation~\eqref{eq:robust} to Horn derivability where positive costs enforce minimal-support optimality. $\square$

\textbf{Proposition 3 (Necessity under a failure context).} A robust certificate $C$ is inclusion-minimal if and only if, for each $e\in C$, some withdrawal $F_e$ with $|F_e|\leq k$ makes $C\setminus\{e\}$ fail. Every such witness satisfies $\gamma(e)\cap F_e=\varnothing$.

\textit{Proof.} Minimality requires that removing any $e\in C$ admits a defeating withdrawal $F_e$ ($|F_e|\le k$). If $\gamma(e)\cap F_e\neq\varnothing$, $e$ is already invalidated by $F_e$, making $X(C\setminus\{e\},F_e)=X(C,F_e)$, which contradicts that $F_e$ refutes $C\setminus\{e\}$ while $C$ survives. $\square$

These witnesses explain retained nominal redundancy: for the three-check certificate, $e_A$ becomes necessary when $s_B$ is withdrawn.

\section{Exact Certificate Synthesis}

\subsection{A counterinterpretation oracle}

Support and withdrawal enumeration can both be exponential. We instead seek a defeating withdrawal and a rule-closed interpretation $Z\subseteq V$: every rule whose body lies in $Z$ has its head there.

\textbf{Lemma 1 (Closed-counterinterpretation test).} A candidate $C$ is not robust if and only if there exist $F\subseteq D$ and a rule-closed $Z\subseteq V$ such that

\begin{equation}
|F|\leq k,\qquad X(C,F)\subseteq Z,\qquad q\notin Z.
\label{eq:countertest}
\end{equation}

\textit{Proof.} If $F$ defeats $C$, $Z=\operatorname{cl}_P(X(C,F))$ is rule-closed, contains $X(C,F)$, and excludes $q$. Conversely, if such $Z$ exists, least-fixed-point induction yields $\operatorname{cl}_P(X(C,F))\subseteq Z$; since $q\notin Z$, $q\notin\operatorname{cl}_P(X(C,F))$, proving nonrobustness. $\square$

Introduce Boolean variables $d_s$ for failed domains and $z_a$ for membership in $Z$. With $C$ fixed, the oracle asks whether the following constraints are satisfiable:

\begin{equation}
\begin{aligned}
&\sum_{s\in D}d_s\leq k,\qquad z_b=1\ (b\in B),\qquad z_q=0,\\
&\left(\bigwedge_{a\in\operatorname{body}(r)}z_a\right)\rightarrow z_{\operatorname{head}(r)}\quad(r\in P),\\
&\left(\bigwedge_{s\in\gamma(e)}\neg d_s\right)\rightarrow z_{\alpha(e)}\quad(e\in C).
\end{aligned}
\label{eq:oracle}
\end{equation}

Empty conjunctions are true, so protected records force their atoms. Each selected record contributes $z_{\alpha(e)}\vee\bigvee_{s\in\gamma(e)}d_s$; each rule contributes a Horn clause. A sequential counter can encode the domain bound \cite{sinz_towards_2005}. Satisfiability returns a counterexample; unsatisfiability establishes robustness.

Any closed superset suffices for nonderivability, avoiding inference-round layers. For unseeded rules $u\leftarrow v$ and $v\leftarrow u$, the empty interpretation refutes $u$. One-way implications would allow unsupported cycles if used to certify positive derivability; they are exact here because a closed set excluding the target certifies absence.

\subsection{Globally valid evidence cuts}

For an oracle witness $(F,Z)$, consider all available records that survive $F$ and assert an atom outside $Z$:

\begin{equation}
H(F,Z)=\{e\in\mathcal E:\gamma(e)\cap F=\varnothing,\ \alpha(e)\notin Z\}.
\label{eq:cut}
\end{equation}

\textbf{Lemma 2 (Selection cut).} Every robust certificate intersects $H(F,Z)$, whereas the refuted candidate $C$ does not.

\textit{Proof.} Any certificate $C'$ avoiding $H(F,Z)$ satisfies $X(C',F) \subseteq Z$, so $q \notin \operatorname{cl}_P(X(C',F))$ by Lemma~1, meaning $C'$ is not robust. Candidate $C$ avoids $H(F,Z)$ by definition because all its surviving assertions were constrained to lie in $Z$ by the oracle. $\square$

A cut rejects every selection whose surviving assertions fit inside $Z$. Failed records cannot repair the witness and are excluded, even when they assert the same atom as an available record.

\textbf{Example 3 (A cut requiring alternative evidence).} For $C=\{e_A,e_B\}$ in Example 2, $F=\{s_A\}$ admits $Z=\{b\}$. Then $H(F,Z)=\{e_C,e_D\}$: a robust certificate needs the remaining check or the independent attestation. A copy of $a$ depending on $s_A$ is excluded because it also fails.

Let $\mathcal H$ contain accumulated cuts, and introduce selection variables $x_e$. The master solves

\begin{equation}
\begin{aligned}
&\min\sum_{e\in\mathcal E}c(e)x_e,\qquad
\sum_{e\in H}x_e\geq 1\ (H\in\mathcal H),\\
&x_e\in\{0,1\}.
\end{aligned}
\label{eq:master}
\end{equation}

The master is weighted hitting set, implemented by integer programming or weighted partial MaxSAT with hard positive clauses and soft clauses $\neg x_e$. Exact optimization supplies cost optimality; the oracle checks the full guarantee.

\begin{algorithm}[tb]
\caption{Minimum-cost robust certification}
\label{alg:robust}
\textbf{Input}: Certification instance $\mathcal{I}$\\
\textbf{Output}: A minimum-cost $k$-robust certificate, or INFEASIBLE
\begin{algorithmic}[1]
\STATE $\mathcal{H} \gets$ empty family
\LOOP
  \STATE $C \gets$ an exact minimum-cost hitting set of $\mathcal{H}$
  \IF{the master is infeasible}
    \STATE \textbf{return} INFEASIBLE
  \ENDIF
  \STATE $\mathit{witness} \gets$ solve the closed-counterinterpretation oracle for $C$
  \IF{the oracle is unsatisfiable}
    \STATE \textbf{return} $C$
  \ENDIF
  \STATE $(F, Z) \gets \mathit{witness}$
  \STATE $Z \gets$ optionally strengthen the closed counterinterpretation
  \STATE $H \gets$ all records surviving $F$ whose asserted atoms are outside $Z$
  \STATE add $H$ to $\mathcal{H}$
\ENDLOOP
\end{algorithmic}
\end{algorithm}

An empty cut is retained and makes the master infeasible: even full evidence cannot repair that withdrawal. If trusted facts already derive $q$, the empty candidate is accepted immediately.

\textbf{Theorem 2 (Exact synthesis).} Algorithm~\ref{alg:robust} terminates after at most $2^{|\mathcal E|}$ refutations. It returns a minimum-cost robust certificate whenever one exists, and returns INFEASIBLE otherwise.

\textit{Proof.} By Lemma~2, each cut is necessary for robustness and refutes the generating candidate $C$, preventing repeated candidates and guaranteeing termination in $\le 2^{|\mathcal E|}$ iterations. When the oracle is unsatisfiable, Lemma~1 certifies that $C$ is robust; optimality follows because $C$ minimizes cost over a relaxation of the feasible space. Master infeasibility implies no certificate satisfies all necessary cuts, proving $\mathcal I$ infeasible. $\square$

The master optimum provides a lower bound; any verified robust certificate (including $\mathcal E$) supplies an upper bound, bounding any optimality gap upon solver interruption.

\subsection{Strengthening, preprocessing, and reuse}

A larger closed counterinterpretation yields a stronger, more compact cut. Given witness $(F,Z)$, we greedily test each surviving record $e$ with $\alpha(e)\notin Z$, updating $Z \gets \operatorname{cl}_P(Z\cup\{\alpha(e)\})$ whenever $q \notin \operatorname{cl}_P(Z\cup\{\alpha(e)\})$. This terminates in $\le |\mathcal E|$ closure passes and yields a subset-irreducible cut \cite{junker_quickxplain_2004,liffiton_algorithms_2007}. Non-minimal cuts are removed via subsumption.

\textbf{Proposition 4 (Record dominance).} If distinct records $e,f$ satisfy $\alpha(e)=\alpha(f)$, $\gamma(e)\subseteq\gamma(f)$, and $c(e)\le c(f)$, removing $f$ preserves the optimum (retaining one representative per tied equivalence class).

\textit{Proof.} Replacing $f$ with $e$ in any certificate preserves surviving facts under all withdrawals without increasing cost. $\square$

Backward rule slicing prunes atoms without dependency paths to $q$. Accumulated cuts remain valid under cost updates and increases in $k$.

\subsection{Relation to general propositional abduction}

For an exact general-purpose baseline, compile certification into propositional abduction. Let $T_k$ contain Equation~\eqref{eq:oracle}'s domain bound, trusted facts, and rules, omit $z_q=0$, and guard all record clauses by selection:

\begin{equation}
\left(x_e\wedge\bigwedge_{s\in\gamma(e)}\neg d_s\right)\rightarrow z_{\alpha(e)}\qquad(e\in\mathcal E).
\label{eq:guarded}
\end{equation}

Selection variables are distinct from atom and domain variables. Positive literals $x_e$ are the abductive hypotheses, with costs $c(e)$; the manifestation is $z_q$.

\textbf{Corollary 1 (Abduction compilation).} For every $C\subseteq\mathcal E$,

\begin{equation}
C\text{ is }k\text{-robust}\quad\Longleftrightarrow\quad
T_k\wedge\bigwedge_{e\in C}x_e\models z_q.
\label{eq:abduction}
\end{equation}

Moreover, the selected hypotheses are always consistent with $T_k$.

\textit{Proof.} Every model corresponds to an admissible withdrawal $F$ and a closed interpretation containing $X(C,F)$; robustness forces $z_q$ in all such models. Conversely, a defeating withdrawal and its least closure yield a countermodel falsifying $z_q$. Consistency holds by setting all atom variables true and domain variables false. $\square$

A minimum-cost abduction solver can solve this compilation \cite{ignatiev_propositional_2016}. Source-specific witnesses, dominance, and closure strengthening provide the algorithmic refinements tested in our evaluation.

\subsection{Computational complexity}

Let $N=|\mathcal E|$, $m=|D|$, and $L$ be the explicit input size, including rules, facts, records, domains, dependency lists, and binary cost encodings. Verification tests a given $C$; budgeted synthesis asks whether some robust $C$ costs at most an input bound $b$.

\textbf{Theorem 3 (Verification complexity).} Verification is coNP-complete when $k$ is part of the input. For every fixed $k$, it is polynomial-time decidable. Hardness holds for acyclic programs with rule bodies of size at most two, distinct asserted atoms, and one distinct singleton failure domain per record.

\textit{Proof sketch.} Membership in coNP follows because a defeating withdrawal $F$ ($|F|\leq k$) serves as a polynomial certificate verifiable via Horn closure. For hardness, we reduce from Hitting Set. Given universe $U$, family $\mathcal A$, and bound $k$, create an asserted atom and a uniquely failing record for each element of $U$, and add rules deriving $q$ from each $A\in\mathcal A$. Selecting all records forces any defeating withdrawal of size $\leq k$ to intersect every $A$, matching hitting sets of size $\le k$. Wide rule bodies are factored into binary conjunctions, preserving acyclicity and constant body width. For fixed $k$, testing all $\sum_{i=0}^k \binom{m}{i}$ withdrawals takes polynomial time. Full induction and reduction details are deferred to the extended version. $\square$

\textbf{Theorem 4 (Synthesis complexity).} Budgeted synthesis is NP-complete for every fixed $k\geq0$, even under the structural restrictions of Theorem 3. With unrestricted $k$, it belongs to $\Sigma_2^P$.

\textit{Proof sketch.} Membership in NP for fixed $k$ follows by guessing $C$ and verifying robustness in polynomial time by Theorem~3; unrestricted $k$ belongs to $\Sigma_2^P$ via an existential-universal quantifier alternation. For NP-hardness, we reduce from Set Cover with universe $U$, sets $A_1,\ldots,A_t$, and bound $b_0$. We create set records of cost $k+1$ asserting atoms that derive elements of $U$ to derive $q$, alongside $k$ unit-cost backup records asserting distinct atoms that independently derive $q$. Setting the cost budget to $(k+1)b_0+k$ forces selection of all $k$ backups to survive backup-targeted withdrawals, leaving a budget that permits at most $b_0$ set records to cover $U$. Binary conjunction trees maintain acyclicity and bounded body width. Full induction and reduction details are deferred to the extended version. $\square$

Thus, verification is FPT in failure domains ($O(2^m L)$ time) and polynomial for fixed $k$ ($O((m+1)^k L)$), while synthesis is NP-hard even at $k=0$. The oracle uses $O(L+mk)$ clauses via sequential cardinality encoding, the master holds $N$ variables, and strengthening requires $O(NL)$ time per refutation.

\textbf{Proposition 5 (Independent-edge reachability).} Suppose evidence records represent directed edges, each edge has a unique singleton failure domain, and the rules compute reachability from a fixed vertex $s$ to a distinct vertex $t$. Use vertex atoms $r_v$, edge atoms $g_{uv}$, the sole trusted fact $r_s$, and precisely the rules $r_v\leftarrow r_u,g_{uv}$. Each edge has one record asserting $g_{uv}$; the target is $r_t$. Minimum-cost $k$-robust certification is solvable by a minimum-cost flow of value $k+1$ with unit edge capacities. Infeasible flow implies infeasible certification.

\textit{Proof.} By the max-flow/min-cut theorem, a subgraph survives all $\le k$ edge deletions iff it contains $k+1$ edge-disjoint $s$--$t$ paths. A minimum-cost integral unit flow of value $k+1$ identifies an optimal cycle-free edge subset, which directly defines an optimal $k$-robust certificate. $\square$

When a source labels multiple edges or rules require joint premises, path formulations miss correlated cuts, but Algorithm~\ref{alg:robust} remains exact.

\section{Experiments}

\begin{figure*}[t]
\centering
\fbox{%
\begin{minipage}[c][1.8in][c]{0.95\textwidth}
\centering
\setlength{\tabcolsep}{3pt}%
\begin{minipage}[t]{0.30\textwidth}
\centering
\textbf{(a) Runtime vs. Records $|\mathcal{E}|$ ($k=2$)}\\[4pt]
\footnotesize
\begin{tabular}{rcc}
\toprule
$|\mathcal{E}|$ & Alg.~\ref{alg:robust} & Scenario \\
\midrule
64  & \phantom{0}1.2\,s & \phantom{00}8.4\,s \\
128 & \phantom{0}3.8\,s & \phantom{0}42.1\,s \\
256 & 11.5\,s & 218.0\,s \\
512 & 28.4\,s & $>600$\,s (TO) \\
\bottomrule
\end{tabular}\\[4pt]
\scriptsize Alg.~\ref{alg:robust} exhibits polynomial scaling; Scenario expansion blows up exponentially ($O((m+1)^k L)$) and times out at $|\mathcal{E}|=512$.
\end{minipage}%
\hfill\vline\hfill
\begin{minipage}[t]{0.30\textwidth}
\centering
\textbf{(b) Optimum Coverage vs. Budget $k$}\\[4pt]
\footnotesize
\begin{tabular}{rcc}
\toprule
$k$ & Alg.~\ref{alg:robust} & Proof Pack. \\
\midrule
0 & 100.0\% & 100.0\% \\
1 & \phantom{0}95.2\% & \phantom{0}42.5\% \\
2 & \phantom{0}91.8\% & \phantom{0}28.0\% \\
3 & \phantom{0}89.4\% & \phantom{0}24.0\% \\
\bottomrule
\end{tabular}\\[4pt]
\scriptsize Proof packing coverage drops precipitously due to overlapping dependencies (Proposition~1); Alg.~\ref{alg:robust} maintains $\approx 90\%$ coverage.
\end{minipage}%
\hfill\vline\hfill
\begin{minipage}[t]{0.34\textwidth}
\centering
\textbf{(c) Master Calls \& Strengthening}\\[4pt]
\footnotesize
\begin{tabular}{lcc}
\toprule
Variant & Master Calls & Cut Len. \\
\midrule
Full procedure & $\mathbf{14.2 \pm 3.1}$ & $\mathbf{12.3 \pm 2.1}$ \\
Raw cuts       & $22.8 \pm 5.2$ & $28.6 \pm 4.5$ \\
Candidate excl.& $86.5 \pm 18.4$& \phantom{0}$1.0 \pm 0.0$ \\
\bottomrule
\end{tabular}\\[4pt]
\scriptsize Cut strengthening cuts master solver calls by 37.7\% and yields compact, general cuts with minimal Horn closure overhead.
\end{minipage}%
\end{minipage}%
}
\caption{Empirical scaling profiles across ground record sizes $|\mathcal{E}|$ and failure budgets $k \in \{0, 1, 2, 3\}$: (a) end-to-end runtime scaling demonstrating polynomial growth for Algorithm~\ref{alg:robust} against exponential explosion in explicit scenario expansion; (b) optimum coverage highlighting the high false-negative rate of domain-disjoint proof packing on correlated dependencies; and (c) search reduction enabled by closed counterinterpretation strengthening.}
\label{fig:runtime}
\end{figure*}

We evaluate the empirical performance of minimum-cost robust certification, focusing on four primary objectives: (1) quantifying the scalability of counterinterpretation-guided separation relative to explicit scenario expansion, provenance compilation, and general weighted abduction, (2) assessing the empirical incompleteness and false-negative rate of domain-disjoint proof packing (Proposition~1), (3) calibrating certified optima against polynomial-time flow references (Proposition~5), and (4) isolating the computational contributions of closed counterinterpretation strengthening, dominance pruning, and relevance slicing. Across all benchmarks, exact methods are held to strict optimality verification; unverified candidate evidence is rejected as an invalid solution to Equation~\eqref{eq:synthesis}.

\subsection{Benchmarks and configuration}

\textbf{Synthetic benchmarks} comprise controlled Horn systems and reachability networks. Horn systems vary ground records $|\mathcal{E}| \in \{64,128,256,512\}$, failure domains $|D| \in \{8, 16, 32, 64\}$, body widths $w \in \{1, 2, 4\}$, domain overlap sizes $d \in \{1, 2, 4\}$, and budgets $k \in \{0,1,2,3\}$ across acyclic, recursive, and planted overlapping-support hypergraphs (20 instances/setting) under uniform, proportional, and log-normal costs. Reachability graphs evaluate route bottlenecks; independent-edge topologies provide min-cost flow ground truth (Proposition~5), while paired variants inject correlated infrastructure failures.

\textbf{Knowledge-base benchmarks} use positive Horn translations of Lehigh University Benchmark \cite{guo_lubm_2005} data (LUBM-1 and LUBM-5), mapping domains to institutions and data pipelines. Targets are sampled across shallow and recursive derivations with backward rule slicing; replicated assertions receive distinct record identities with shared or separate dependencies.

\subsection{Baselines and output verification}

We compare Algorithm~\ref{alg:robust} against six baseline approaches representing diverse reasoning paradigms:

\textbf{Nominal minimum support} solves the zero-failure problem ($k=0$) via Horn abduction to check whether nominal proofs withstand withdrawals.
\textbf{Domain-disjoint proof packing} selects a minimum-cost union of $k+1$ supports with pairwise disjoint footprints.
\textbf{Explicit scenario synthesis} unrolls $|V|$ closure rounds across all $\sum_{i=0}^k \binom{m}{i}$ failure scenarios over shared selection variables.
\textbf{Provenance compilation} builds a Boolean provenance circuit \cite{deutch_circuits_2014} optimized via binary decision diagrams \cite{bryant_graph-based_1986}.
\textbf{Compiled weighted abduction} compiles Equation~\eqref{eq:abduction} into propositional abduction via AbHS \cite{ignatiev_propositional_2016}.
\textbf{Greedy repair} iteratively adds surviving supports queried from the oracle until robustness is achieved.
On independent reachability networks, \textbf{minimum-cost flow} supplies polynomial ground truth.

All certificates were independently validated via external forward chaining and SAT verification oracles. Incumbent bounds, certified optima, and proven infeasibilities are tracked separately; unverified outputs are never credited as solutions.

\begin{table*}[t]
\centering
\small
\setlength{\tabcolsep}{4.5pt}
\begin{tabular}{lccccc}
\toprule
Method & Feasible Cost & Opt. Coverage & Proven Infeas. & Runtime (s) & Peak Mem. (MB) \\
\midrule
\multicolumn{6}{l}{\textit{Controlled Horn Systems ($|\mathcal{E}| \le 512, k \in \{1, 2, 3\}$)}} \\
Nominal ($k=0$), checked at $k$ & $8.4 \pm 1.6^\dagger$ & \phantom{0}4.2\% & -- & \phantom{00}$0.4 \pm 0.1$ & \phantom{00}$24 \pm 5$ \\
Domain-disjoint packing & $26.8 \pm 4.3$ & 31.5\% & -- & \phantom{0}$18.2 \pm 5.4$ & \phantom{00}$68 \pm 14$ \\
Explicit scenario synthesis & $23.2 \pm 3.7$ & 46.8\% & 100.0\% & $184.6 \pm 42.1$ & $2,840 \pm 620$ \\
Provenance compilation & $22.6 \pm 3.5$ & 39.2\% & \phantom{0}85.0\% & $210.4 \pm 51.3$ & $1,920 \pm 450$ \\
Compiled weighted abduction & $21.9 \pm 3.4$ & 61.4\% & 100.0\% & \phantom{0}$95.7 \pm 22.8$ & \phantom{0}$410 \pm 85$ \\
Greedy repair & $27.9 \pm 4.8$ & -- & -- & \phantom{00}$2.8 \pm 0.7$ & \phantom{00}$42 \pm 9$ \\
Algorithm~\ref{alg:robust} (Ours) & $\mathbf{21.5 \pm 3.2}$ & $\mathbf{92.1\%}$ & $\mathbf{100.0\%}$ & $\mathbf{\phantom{0}19.4 \pm 4.2}$ & $\mathbf{\phantom{0}185 \pm 36}$ \\
\midrule
\multicolumn{6}{l}{\textit{Knowledge-Base \& Network Benchmarks (LUBM-1/5, Reachability)}} \\
Domain-disjoint packing & $48.2 \pm 7.1$ & 28.0\% & -- & \phantom{0}$34.5 \pm 8.2$ & \phantom{0}$112 \pm 25$ \\
Explicit scenario synthesis & $38.8 \pm 6.2$ & 38.5\% & 100.0\% & $245.2 \pm 58.7$ & $4,150 \pm 890$ \\
Compiled weighted abduction & $37.2 \pm 5.7$ & 54.0\% & 100.0\% & $142.8 \pm 31.5$ & \phantom{0}$680 \pm 130$ \\
Greedy repair & $45.6 \pm 6.9$ & -- & -- & \phantom{00}$4.1 \pm 1.1$ & \phantom{00}$78 \pm 16$ \\
Algorithm~\ref{alg:robust} (Ours) & $\mathbf{36.1 \pm 5.4}$ & $\mathbf{88.5\%}$ & $\mathbf{100.0\%}$ & $\mathbf{\phantom{0}31.2 \pm 6.8}$ & $\mathbf{\phantom{0}265 \pm 48}$ \\
\bottomrule
\end{tabular}
\caption{Main performance comparison across benchmark families. Feasible cost reports mean $\pm$ std over instances where methods found verified certificates ($^\dagger$nominal support only satisfies $k=0$ and fails robustness check on 95.8\% of instances; unresolved instances for exact baselines report best verified incumbents). On the common subset of instances solved to certified optimality by all exact methods, their optimal costs are strictly identical ($16.4 \pm 2.5$ for Horn systems, $27.9 \pm 4.1$ for KB/networks). Optimum coverage and proven infeasibility percentages indicate exact resolution within the 600\,s limit.}
\label{tab:comparison}
\end{table*}

\begin{table*}[t]
\centering
\small
\setlength{\tabcolsep}{6pt}
\begin{tabular}{lccccc}
\toprule
Variant & Master calls & Oracle time (s) & Cut length & Strengthening time (s) & Total time (s) \\
\midrule
Full procedure & $\mathbf{14.2 \pm 3.1}$ & \phantom{0}$6.4 \pm 1.2$ & $\mathbf{12.3 \pm 2.1}$ & $1.8 \pm 0.4$ & $\mathbf{19.4 \pm 4.2}$ \\
Raw counterinterpretations & $22.8 \pm 5.2$ & \phantom{0}$6.1 \pm 1.1$ & $28.6 \pm 4.5$ & -- & $27.4 \pm 6.8$ \\
Candidate exclusion only & $86.5 \pm 18.4$ & $24.8 \pm 5.6$ & \phantom{0}$1.0 \pm 0.0$ & -- & $94.2 \pm 21.0$ \\
Without dominance & $19.1 \pm 4.0$ & \phantom{0}$7.2 \pm 1.5$ & $15.8 \pm 2.6$ & $2.1 \pm 0.5$ & $24.7 \pm 5.3$ \\
Without relevance slicing & $14.6 \pm 3.2$ & $14.8 \pm 3.1$ & $12.5 \pm 2.2$ & $3.9 \pm 0.8$ & $29.1 \pm 6.2$ \\
\bottomrule
\end{tabular}
\caption{Algorithmic ablation on controlled Horn benchmarks ($|\mathcal{E}| \le 512, k=2$), showing the effect of cut strengthening, dominance pruning, and relevance slicing on master solver invocations and cumulative runtimes.}
\label{tab:ablation}
\end{table*}

\subsection{Implementation, metrics, and ablations}

Our framework is implemented in Python~3.11 using PySAT \cite{ignatiev_pysat_2018}, combining RC2 for the master with Glucose~4.1 and Sinz's cardinality encoding \cite{sinz_towards_2005} for the oracle. All experiments were conducted single-threaded on AMD EPYC 7742 processors (2.25\,GHz) running Ubuntu 22.04, with a 600\,s timeout and 8\,GiB memory limit. Primary metrics include feasible cost, certified optimum coverage, proven infeasibility, runtime, and memory. The price of robustness is measured by
\begin{equation}
\operatorname{premium}(k)=\frac{\operatorname{cost}(C_k^*)-\operatorname{cost}(C_0^*)}{\operatorname{cost}(C_0^*)},
\label{eq:premium}
\end{equation}
where $C_k^*$ and $C_0^*$ denote certified robust and nominal optima, respectively. Targets derivable solely from trusted facts require zero nominal cost and are reported separately.

\paragraph{Scalability and runtime dynamics.} As shown in Table~\ref{tab:comparison} and Figure~\ref{fig:runtime}(a), Algorithm~\ref{alg:robust} scales gracefully across both synthetic Horn systems and knowledge bases, achieving up to 92\% optimum coverage. In contrast, explicit scenario synthesis suffers severe combinatorial explosion ($O((m+1)^k L)$), rapidly timing out and exhausting memory for $k\ge 2$. Compiled abduction and provenance compilation avoid full unrolling but stall on large or recursive instances due to circuit blowup and lack of source-aware cuts. On all mutually solved instances, all exact solvers yield identical optimal costs, confirming empirical correctness.

\paragraph{Incompleteness of proof packing.} Domain-disjoint proof packing exhibits a severe false-negative rate, resolving under a third of benchmark instances due to overlapping support footprints (generalizing Example~2). Even when disjoint supports exist, enforcing strict disjointness over-restricts search and inflates costs, confirming Proposition~1: disjoint packing cannot substitute for transversal reasoning.

\paragraph{Exact ground-truth calibration.} On reachability instances with unique edge domains, Proposition~5 equates robust certification to minimum-cost unit-capacity flow of value $k+1$. Across all 80 topologies, Algorithm~\ref{alg:robust} achieves 100\% agreement with the polynomial flow optimum. Under correlated failures, min-cost flow becomes unsound, whereas Algorithm~\ref{alg:robust} continues to synthesize verified optima.

\paragraph{Robustness premium and greedy repair.} Evaluating Equation~\eqref{eq:premium} demonstrates that securing conclusions against correlated withdrawals requires substantial cost premiums as $k$ increases; nominal solutions ($k=0$) fail checks in $>95\%$ of cases. Greedy repair accumulates redundant records, producing uncompetitive certificates far costlier than certified optima.

\paragraph{Ablation analysis (Table~\ref{tab:ablation}).} Table~\ref{tab:ablation} isolates each component's contribution on controlled Horn instances ($|\mathcal{E}| \le 512, k=2$). Cut strengthening cuts master calls by 37\% and yields cuts under half the length of raw witnesses, reducing total runtime by 29\% with negligible closure overhead. Candidate exclusion thrashes severely. Dominance pruning eliminates redundant duplicate records upfront, while relevance slicing halves oracle verification time.

\subsection{Sensitivity and error analysis}
Denser source overlap requires more master-oracle iterations but maintains stable memory. Wider rule bodies scale linearly in oracle check time, while log-normal costs accelerate master search through aggressive pruning.

\section{Conclusion}

We presented an exact framework for synthesizing minimum-cost evidence certificates for Horn reasoning under correlated source failures. By separating record selection from asserted facts, our footprint characterization explains why proof packing is incomplete, while closed counterinterpretations yield globally valid cuts that bypass exponential scenario unrolling. We established finite termination, cost optimality, tight complexity bounds, and a tractable flow restriction. Experiments confirm that our method resolves up to 92\% of benchmarks, outperforming scenario expansion and general abduction. Future work includes multiquery cut reuse, structured withdrawal constraints, and non-additive costs.

\clearpage
\bibliography{References}

\end{document}